\documentclass[runningheads]{llncs}

\usepackage{accv}
\usepackage{accvabbrv}

\usepackage{graphicx}
\usepackage{booktabs}

\usepackage[accsupp]{axessibility}  %

\usepackage{hyperref}
\usepackage{orcidlink}
\usepackage{amsfonts}
\usepackage{booktabs}
\usepackage{multirow}
\usepackage{makecell}
\usepackage{graphicx}
\usepackage{subcaption}
\usepackage[font=small]{caption}
\usepackage{algorithm}
\usepackage{algpseudocode}
\usepackage{cite}
\usepackage{censor}
\usepackage{xcolor}

\usepackage[acronym]{glossaries}
\glsdisablehyper

\newacronym{sg}{3DSGs}{3D Scene Graphs}
\newacronym{nerf}{NeRF}{Neural Radiance Field}
\newacronym{gs}{3D-GS}{3D Gaussian Splatting}
\newacronym{vlm}{VLM}{Vision-Language Models}
\newacronym{llm}{LLM}{Large Language Models}
\newacronym{slam}{SLAM}{Simultaneous Localization and Mapping}
\newacronym{tsdf}{TSDF}{Truncated Signed Distance Function}
\newcommand{\coolname}{\text{OpenSplatGraph}}

\begin{document}

\title{OpenSplatGraph: \\
From Dense Semantic Maps to Structured Scene Graphs for Open-Vocabulary Robot Perception} 

\titlerunning{OpenSplatGraph}

\author{Binh Long Nguyen\inst{1,2}\orcidlink{0009-0001-7564-2476} \and
Kien Nguyen\inst{1}\orcidlink{0000-0002-3466-9218} \and \\
Clinton Fookes\inst{1}\orcidlink{0000-0002-8515-6324} \and
Peyman Moghadam \inst{1,2}\orcidlink{0000-0002-8169-3560}
}

\authorrunning{B. L. Nguyen et al.}

\institute{School of Electrical Engineering and Robotics,
Queensland University of Technology (QUT), Brisbane, QLD 4000, Australia
\email{\{binhlong.nguyen, k.nguyenthanh, c.fookes, peyman.moghadam\}@qut.edu.au}\\
\and
CSIRO Robotics, CSIRO, Brisbane, QLD 4069, Australia\\
\email{\{binhlong.nguyen, peyman.moghadam\}@csiro.au}}

\maketitle

\begin{abstract}
Dense 3D mapping with semantic understanding is essential for robotic perception in complex environments. Recent 3D Gaussian Splatting-based mapping approaches enable high-fidelity geometry and efficient open-vocabulary perception, but typically represent semantics as unstructured feature fields that limit object-centric reasoning. In contrast, 3D scene graphs explicitly model objects and their relationships for structured reasoning, but are commonly constructed from sparse geometric representations that do not fully exploit dense semantic maps.
In this work, we present \coolname{}, a unified framework that constructs persistent 3D scene graphs directly from an online Gaussian-based open-vocabulary semantic map. The proposed framework augments the dense semantic map with a reliability-aware semantic field that maintains lightweight observation statistics for confidence-aware, query-conditioned object extraction. Extracted object instances are associated with persistent graph nodes, allowing object attributes and relationships to be incrementally updated across observations and queries. By tightly coupling dense semantic mapping with persistent object-centric representations, our framework supports both language-guided object grounding and structured relational reasoning while preserving the geometric fidelity of Gaussian-based mapping.
Comprehensive evaluations on standard 3D scene understanding benchmarks and real-world robotic experiments demonstrate that \coolname{} achieves competitive performance for online open-vocabulary perception and downstream robotic tasks.
Project page: \url{https://csiro-robotics.github.io/OpenSplatGraph}.

\end{abstract}

\section{Introduction}
\label{sec:intro}

Dense 3D mapping with semantic understanding is fundamental to robotic perception, supporting navigation, manipulation, and interaction in complex environments~\cite{huang2023visual,jatavallabhula2023conceptfusion,zhou2024navgpt}. Recent advances in vision-language foundation models~\cite{radford2021learning,li2022blip} have further enabled open-vocabulary 3D scene understanding that responds to natural language queries and generalizes to previously unseen objects.

Meanwhile, \gls{gs}~\cite{kerbl20233d} has established an efficient representation for high-fidelity 3D reconstruction with real-time rendering. Online GS-based mapping systems~\cite{keetha2024splatam,matsuki2024gaussian,ha2024rgbd} integrate Gaussian primitives with \gls{slam} for geometrically consistent mapping. Building on this representation, open-vocabulary approaches~\cite{qin2024langsplat,wu2024opengaussian,yang2025opengs,nguyen2026ilov3splat} incorporate vision-language features into dense Gaussian maps, enabling language-guided retrieval of scene regions. However, semantics are often represented as unstructured per-voxel or per-Gaussian feature fields rather than explicit object-level entities. Consequently, they provide limited support for object-centric reasoning and modeling inter-object relationships.

In parallel, \gls{sg} address this limitation by representing environments as graphs whose nodes correspond to objects and whose edges encode semantic or spatial relationships~\cite{armeni20193d}. Recent advances have extended scene graphs to online incremental systems~\cite{wu2021scenegraphfusion,hughes2022hydra}, open-vocabulary reasoning~\cite{gu2024conceptgraphs,koch2024open3dsg,maggio2024clio}, and large-scale robotic applications~\cite{deng2024opengraph,yin2024sg}. Most open-vocabulary scene graph methods construct graphs from RGB-D observations by projecting 2D detections into point clouds or meshes. While these approaches provide compact and interpretable scene abstractions for reasoning, they typically rely on sparse geometric representations that cannot fully exploit the dense geometric and semantic information available in Gaussian-based maps.

\begin{figure}[t]
	\centering
    \includegraphics[width=0.9\columnwidth]{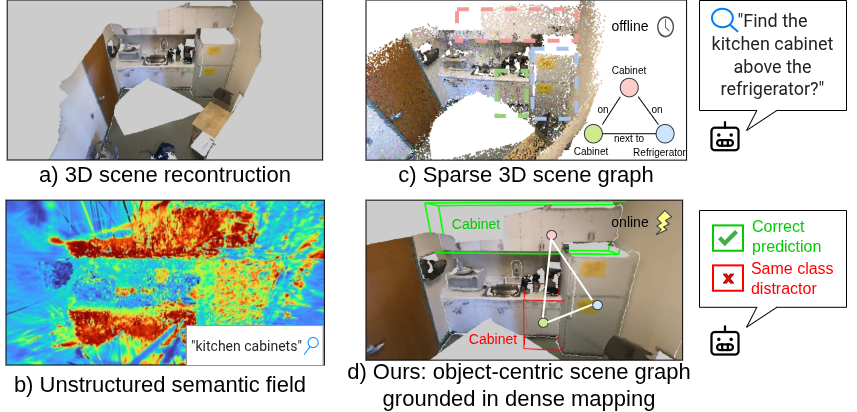}
	\centering
    \caption{
    Comparison of complementary representations for open-vocabulary 3D scene understanding. Dense semantic maps provide high-fidelity geometry and language-aligned features but represent semantics as unstructured feature fields. Scene graphs explicitly model objects and their relationships for structured reasoning, but typically rely on sparse geometric representations. Our framework bridges these paradigms by constructing a persistent 3D scene graph directly from an online dense semantic map.
    }
	\label{fig:comparison}
\end{figure} 

As illustrated in Fig.~\ref{fig:comparison}, robotic systems benefit from these two complementary scene representations.
Dense geometric-semantic maps provide high-fidelity 3D reconstruction for language-grounded perception and localization, whereas object-centric scene graphs explicitly model objects and their relationships for structured reasoning and task planning. However, existing approaches typically emphasize one representation over the other, making it difficult to simultaneously achieve both high-fidelity mapping and structured scene understanding.

In this paper, we propose \coolname{}, a framework that bridges these two paradigms by grounding 3D scene graphs directly in an online open-vocabulary \gls{gs} semantic map. Rather than treating the dense semantic field and the scene graph as alternative representations, we maintain both throughout operation, allowing each to serve a complementary role. 
The dense semantic map preserves fine-grained geometric and semantic information for open-vocabulary perception, while the scene graph incrementally accumulates persistent object-centric knowledge for structured reasoning.
To support object construction, we introduce a reliability-aware semantic field that augments the dense semantic map with lightweight observation statistics, enabling confidence-aware ranking of query-conditioned object instances.
Extracted object instances are assigned persistent identities and incrementally incorporated into a 3D scene graph, allowing object attributes and relationships to be consistently accumulated across multiple queries.
By jointly maintaining these two representations, \coolname{} preserves the geometric fidelity of dense mapping while progressively constructing an interpretable object-centric representation for language grounding, relational reasoning, and downstream robotic tasks.

Our contributions are summarized as follows:
\begin{itemize}
    \item We introduce \coolname{}, a unified framework that constructs persistent 3D scene graphs directly from online open-vocabulary Gaussian semantic maps, enabling high-fidelity dense mapping and structured scene understanding within a single system.

    \item We propose a reliability-aware semantic field that leverages lightweight per-voxel observation statistics for robust query-conditioned object extraction under partial observations and semantic ambiguity.

    \item We design a scene graph construction strategy that associates extracted object instances with persistent identities and incrementally accumulates object-centric knowledge across queries for structured relational reasoning.

    \item We validate the framework on standard 3D scene understanding benchmarks and real-world robotic experiments, demonstrating competitive performance for online open-vocabulary perception and downstream robotic tasks.
\end{itemize}

\section{Related Work}
\label{sec:related}

\subsection{Open-Vocabulary 3D Scene Understanding}

Vision-language foundation models~\cite{radford2021learning,li2022blip} have enabled open-vocabulary 3D scene understanding by aligning visual representations with natural language. Early methods, such as PointCLIP~\cite{zhang2022pointclip} and ConceptFusion~\cite{jatavallabhula2023conceptfusion}, leverage language features for point cloud understanding, while LERF~\cite{kerr2023lerf} extends language-guided retrieval to \gls{nerf} representations.
More recently, \gls{gs}~\cite{kerbl20233d} has emerged as an efficient representation for open-vocabulary 3D perception and robotic applications~\cite{nguyen2026ilov3splat,zielinski2026neo}. Representative methods, including LangSplat~\cite{qin2024langsplat}, LEGaussian~\cite{shi2024language}, OpenGaussian~\cite{wu2024opengaussian}, and InstanceGaussian~\cite{li2025instancegaussian}, associate language features with Gaussian primitives for open-vocabulary querying. However, these methods primarily represent semantics as unstructured feature fields, limiting their support for object-centric and relational reasoning.

\subsection{3D Gaussian-Based Scene Mapping}

\gls{gs} has been integrated into \gls{slam} systems for efficient online dense mapping and high-fidelity rendering. Methods such as SplaTAM~\cite{keetha2024splatam}, MonoGS~\cite{matsuki2024gaussian}, and GS-ICP SLAM~\cite{ha2024rgbd} jointly optimize Gaussian representations and camera poses for real-time reconstruction. Subsequent works incorporate semantic information through supervised labels~\cite{li2024sgs,zhu2025semgauss} or open-vocabulary features, as demonstrated by OpenGS-Fusion~\cite{yang2025opengs}. 
These approaches provide geometrically accurate and semantically rich maps for robotic perception.
However, they primarily focus on online reconstruction and semantic fusion, without explicitly maintaining persistent object-centric representations for structured reasoning and task planning.

\subsection{3D Scene Graphs}

\gls{sg}~\cite{armeni20193d} represent objects as nodes and their relationships as edges, providing interpretable scene abstractions for planning and human-robot interaction~\cite{rana2023sayplan,werby2024hierarchical}. 
Recent advances~\cite{catalano20253d} highlight rapid progress from online incremental graph construction~\cite{hughes2022hydra,wu2021scenegraphfusion} to open-vocabulary scene understanding through ConceptGraphs~\cite{gu2024conceptgraphs}, Open3DSG~\cite{koch2024open3dsg}, and Clio~\cite{maggio2024clio}, and large-scale robotic applications~\cite{deng2024opengraph,yin2024sg}. 
These methods typically construct graphs from point clouds or meshes, providing compact object-centric representations for downstream reasoning. Despite these advances, many open-vocabulary approaches infer object relationships from 2D observations using \gls{vlm}, resulting in graph representations that are weakly grounded in dense 3D geometry.

Recent studies~\cite{ge2025dynamicgsg,wang2025gaussiangraph} have explored grounding scene graphs directly in Gaussian representations. Our work follows this emerging direction but adopts a different architectural perspective. Rather than treating the dense maps and the scene graph as alternative world models, we explicitly maintain both throughout operation. The dense semantic field serves as the geometric and semantic foundation for reliable object extraction, while the scene graph incrementally accumulates object-centric knowledge for structured and relational reasoning.

\section{Methodology}
\label{sec:method}

\begin{figure*}[t]
	\centering
	\includegraphics[width=1.0\textwidth]{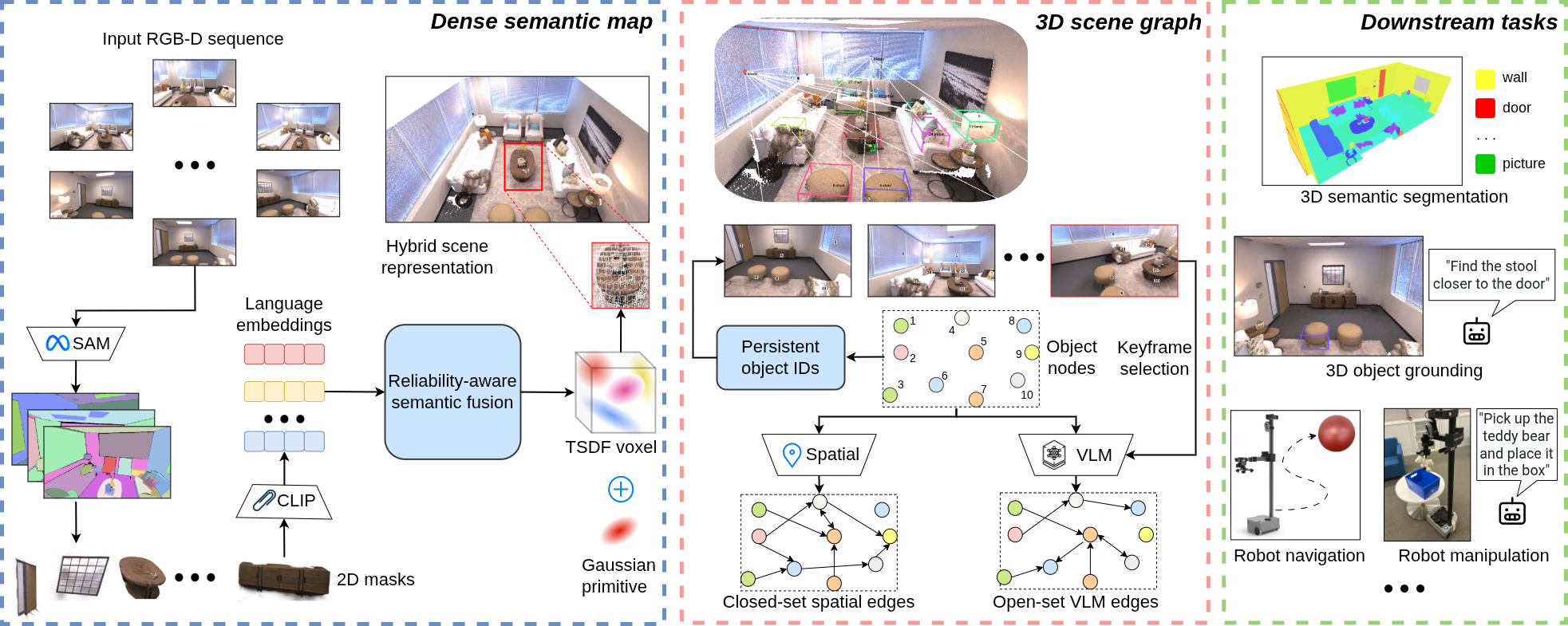}
	\centering
    \caption{Overview of \coolname{}. 
    Starting from a multi-view RGB-D sequence, the framework incrementally constructs an online open-vocabulary dense semantic map together with a reliability-aware semantic field for robust query-conditioned object extraction. Extracted object instances are then associated across queries and incrementally accumulated into a persistent 3D scene graph. By jointly maintaining the dense semantic field and the scene graph, \coolname{} enables structured relational reasoning while preserving high-fidelity geometric and semantic information for downstream robotic tasks.}
	\label{fig:overview}
\end{figure*}

We propose \coolname{}, a unified framework that jointly maintains a dense open-vocabulary semantic field and a persistent 3D scene graph throughout online operation. The dense semantic field serves as a geometrically consistent foundation for query-conditioned object extraction, while the scene graph incrementally accumulates object-centric knowledge to support structured reasoning across multiple queries. Rather than maintaining a fixed object inventory, \coolname{} extracts objects on demand from language queries by grouping spatially and semantically consistent regions in the semantic field. Extracted instances are associated with persistent graph nodes and enriched with semantic and relational information as the map evolves. In this way, \coolname{} preserves the fidelity of dense semantic mapping while progressively constructing an interpretable scene representation for downstream robotic tasks. An overview of the framework is shown in Fig.~\ref{fig:overview}.

\subsection{Reliability-Aware Semantic Field}
\label{sec:method:semantic}

The dense semantic field forms the perception layer of \coolname{}, providing a geometrically consistent representation for query-conditioned object extraction. Built upon an online open-vocabulary Gaussian semantic map, the field maintains fused semantic features together with lightweight reliability statistics that characterize the quality of semantic observations. These statistics regulate semantic fusion and assess the reliability of extracted object candidates, improving robustness under partial visibility, occlusions, and semantic ambiguity.

\subsubsection{Semantic Representation.}
\label{sec:semantic:substrate}

\coolname{} adopts a hybrid semantic mapping framework~\cite{yang2025opengs}, combining voxel-based \gls{tsdf} geometry with 3D Gaussian primitives. The \gls{tsdf} supports incremental geometric and semantic fusion, while the Gaussians capture fine scene details and enable high-fidelity rendering. Given posed RGB-D observations, we extract 2D instance masks using SAM~\cite{kirillov2023segment} and associate them with language-aligned CLIP embeddings~\cite{radford2021learning}. These observations are incrementally integrated into a semantic field that remains consistent across viewpoints and over time.

Formally, the hybrid map is represented by a sparse voxel set $\mathcal{V}=\{v_j\}_{j=1}^{V}$, where each voxel $v$ is associated with a set of Gaussian primitives $\mathcal{G}_v$. Let $d$ denote the semantic embedding dimension (e.g., $d=512$ for CLIP). 
Each voxel stores a fused semantic embedding $F_v\in\mathbb{R}^d$, an observation count $c(v)$, an accumulated semantic confidence $w(v)$, and a 3D coordinate $X_v\in\mathbb{R}^3$ in the global reference frame.
For each observed voxel, an incoming semantic embedding $f_v$ with confidence $\omega_v$ is integrated as:
\begin{equation}
    F_v \leftarrow
    \frac{c(v)F_v+\omega_v f_v}{c(v)+\omega_v}.
    \label{eq:semantic_fusion}
\end{equation}
Here, $c(v)$ controls the contribution of the stored embedding based on the number of prior observations, while $\omega_v$ weights the incoming embedding according to its observation confidence. This mixed weighting scheme reduces the influence of low-confidence observations on the accumulated embedding.
After fusion, we update $c(v)\leftarrow c(v)+1$ and $w(v)\leftarrow w(v)+\omega_v$, where the accumulated confidence $w(v)$ is used for subsequent instance scoring.
Given a language query $q$, we obtain its normalized text embedding $e(q)\in\mathbb{R}^d$ and compute the voxel-level similarity $s_q(v)=\cos(e(q),F_v)$. This results in a dense semantic foundation that supports query-conditioned object extraction and subsequent scene graph construction.

\subsubsection{Reliability-Aware Semantic Modeling.}
\label{sec:semantic:reliability}

Semantic fusion over long RGB-D sequences naturally produces observations of varying reliability due to occlusions, partial visibility, and semantic ambiguity. To capture this uncertainty, \coolname{}
augments the semantic field with a per-voxel residual semantic
disagreement $r(v)$. Together with the observation count $c(v)$
and accumulated confidence $w(v)$, it forms the per-voxel
statistics:
\begin{equation}
    \mathcal{S}
    =
    \left\{
    (c(v),w(v),r(v))
    \mid v\in\mathcal{V}
    \right\}.
\end{equation}
For each incoming observation, $r(v)$ is updated using an
exponential moving average of the normalized cosine disagreement
between the incoming and previously fused semantic embeddings,
computed before semantic fusion. The residual is initialized
to zero, with a fixed decay factor used throughout the sequence.

Given a candidate object region represented by a voxel subset $\mathcal{V}_k\subset\mathcal{V}$, we define an instance confidence score $C(\mathcal{V}_k)\in[0,1]$ that favors well-observed and
semantically consistent regions:
\begin{equation}
\begin{aligned}
C(\mathcal{V}_k) =\;& \lambda \cdot 
\tanh\!\Bigl(
\tfrac{1}{|\mathcal{V}_k|}
\sum_{v\in\mathcal{V}_k} \tfrac{w(v)}{\kappa}
\Bigr)
+ (1-\lambda)\cdot 
\mathrm{sat}_{[0,1]}\!\Bigl(
1 - \tfrac{1}{|\mathcal{V}_k|}
\sum_{v\in\mathcal{V}_k} r(v)
\Bigr),
\end{aligned}
\label{eq:instance_conf}
\end{equation}
where $\lambda\in(0,1)$ controls the trade-off between observation strength and semantic stability, $\kappa>0$ is a normalization constant, and $\mathrm{sat}_{[0,1]}(x)=\min(1,\max(0,x))$.
The confidence score is used solely for ranking and filtering candidate object instances and requires no supervised calibration.

\subsubsection{Query-conditioned Object Extraction.}
\label{sec:semantic:instances}

Given a language query $q$, we extract a set of object instances $\mathcal{O}_q=\{O_1,\ldots,O_K\}$ from the voxel-level similarity field $s_q(v)$. Each instance corresponds to a spatially coherent voxel cluster with associated semantic and geometric attributes. 
We first threshold the similarity field using $m_q(v)=\mathbb{I}[s_q(v)>\delta]$ to obtain the corresponding voxel coordinates $X_q=\{X_v\mid m_q(v)=1\}$.
We then cluster $X_q$ in Euclidean space using DBSCAN with neighborhood radius $r$ and minimum number of points $k_{\min}$. The radius is set proportional to the voxel size to ensure scale consistency across maps.

For each resulting voxel cluster $\mathcal{V}_k$, we compute its centroid $\mu_k=\frac{1}{|\mathcal{V}_k|} \sum_{v\in\mathcal{V}_k}X_v$, axis-aligned bounding box (AABB), instance score $S_k$ as the mean voxel-level similarity, and reliability-based confidence $C(\mathcal{V}_k)$ from Eq.~\eqref{eq:instance_conf}.
Candidate instances are ranked by a weighted combination of $S_k$ and $C(\mathcal{V}_k)$, prioritizing semantic similarity and using cluster size as a tie-breaker. The top-ranked candidates are retained for subsequent processing. The complete extraction procedure is summarized in Alg.~\ref{alg:instance_extraction}.
Importantly, object extraction is conditioned on the language query, whereas the underlying semantic field remains persistent and is incrementally updated during mapping.

\begin{algorithm}[t]
\caption{Query-conditioned object extraction from voxel similarity scores.}
\label{alg:instance_extraction}
\begin{algorithmic}[1]
\Require Query \(q\); voxel embeddings \(F\); voxel coordinates \(X\); voxel stats \(\mathcal{S}\); threshold \(\delta\); clustering params \((r, k_{\min})\); top-\(K\)
\Ensure Query-conditioned instances \(\{O_k\}\) with voxel set \(\mathcal{V}_k\), centroid \(\mu_k\), AABB, score \(S_k\), confidence \(C_k\)
\State \(e \gets \mathrm{EncodeText}(q)\) \Comment{normalized text embedding}
\For{\(v = 1\) to \(V\)}
  \State \(s(v) \gets \cos(e, F_v)\) 
\EndFor
\State \(\mathcal{M} \gets \{\, v \mid s(v) > \delta \,\}\)
\If{\(\mathcal{M} = \emptyset\)}
  \State \Return \(\emptyset\)
\EndIf
\State \(\mathcal{C} \gets \mathrm{DBSCAN}(\{X_v\}_{v\in\mathcal{M}};\ r,\ k_{\min})\)
\Comment{a set of clusters}
\ForAll{cluster \(\mathcal{V}_k \in \mathcal{C}\)}
  \State \(\mu_k \gets \mathrm{mean}\{X_v \mid v\in\mathcal{V}_k\}\)
  \State \(\mathrm{AABB}_k \gets \bigl(\min_{v\in\mathcal{V}_k} X_v,\ \max_{v\in\mathcal{V}_k} X_v\bigr)\)
  \State \(S_k \gets \mathrm{mean}\{s(v) \mid v\in\mathcal{V}_k\}\)
  \State \(C_k \gets \mathrm{InstanceConfidence}(\mathcal{V}_k;\mathcal{S})\) \Comment{computed via Eq.~\eqref{eq:instance_conf}}
  \State \(O_k \gets (\mathcal{V}_k,\mu_k,\mathrm{AABB}_k,S_k,C_k)\)
\EndFor
\State \(\{O_k\} \gets \mathrm{RankAndSelect}(\{O_k\}, K)\)
\State \Return \(\{O_k\}\)
\end{algorithmic}
\end{algorithm}

\subsection{Persistent 3D Scene Graph Construction}
\label{sec:method:scenegraph}

The persistent scene graph forms the reasoning layer of \coolname{}, incrementally accumulating object-centric knowledge extracted from the reliability-aware semantic field. 
Rather than reconstructing a scene graph for every query, confidence-ranked object instances are associated with existing graph nodes whenever possible, enabling incremental updates to node attributes and inter-object relations.
The resulting graph provides a compact and interpretable representation for structured reasoning while preserving the underlying dense semantic field for perception. We represent the scene as a graph $G=(\mathcal{N},\mathcal{E})$, where nodes $\mathcal{N}$ define object instances and edges $\mathcal{E}$ encode inter-object relations.

\subsubsection{Object Association.}
\label{sec:semantic:ids}

To maintain persistent object identities during mapping, \coolname{} maintains a set of previously extracted object instances and associates newly extracted objects with this set before updating the scene graph.

Let $\hat{\mathcal{O}}=\{\hat{O}_j\}$ denote the set of previously extracted object instances, each associated with a persistent identifier (ID), semantic label, and centroid. For each newly extracted instance $O_i$ with label $\ell_i$ and centroid $\mu_i$, candidate matches are restricted to instances in $\hat{\mathcal{O}}$ with the same semantic label. Among these candidates, the nearest unmatched centroid within a distance threshold $\tau$ is selected. If a valid match is found, its persistent identifier is reused; otherwise, a new identifier is assigned. This one-to-one association preserves object identities across extraction steps and supports incremental graph updates. The complete association procedure is summarized in Alg.~\ref{alg:persistent_id}.

\begin{algorithm}[t]
\caption{Persistent object association via label-aware centroid matching.}
\label{alg:persistent_id}
\begin{algorithmic}[1]
\Require New instances $\{O_i\}$ with label $\ell_i$ and centroid $\mu_i$;
         previous instances $\{\hat{O}_j\}$ with identifier
         $\hat{\mathrm{id}}_j$, label $\hat{\ell}_j$, centroid $\hat{\mu}_j$;
         distance threshold $\tau$
\Ensure Assigned persistent IDs for all $O_i$
\State $\mathrm{Prev}[\ell] \gets
    \{\hat{O}_j \mid \hat{\ell}_j = \ell\}$
\Comment{group previous instances by label}
\State Mark all previous IDs as unused
\ForAll{new instance $O_i$}
    \State $\mathrm{Candidates} \gets
        \{\hat{O}_j \in \mathrm{Prev}[\ell_i]
        \mid \hat{\mathrm{id}}_j \text{ unused},
        \ \|\mu_i-\hat{\mu}_j\|_2 \le \tau\}$

    \If{$\mathrm{Candidates} \neq \emptyset$}
        \State $j^\star \gets
            \arg\min_{\hat{O}_j \in \mathrm{Candidates}}
            \|\mu_i-\hat{\mu}_j\|_2$
        \State $O_i.\mathrm{id} \gets
            \hat{\mathrm{id}}_{j^\star}$
        \State Mark $\hat{\mathrm{id}}_{j^\star}$ as used
    \Else
        \State $O_i.\mathrm{id} \gets \mathrm{NewID}()$
    \EndIf
\EndFor
\State \Return $\{O_i\}$
\end{algorithmic}
\end{algorithm}

\subsubsection{Nodes.} 
Each graph node $N_i \in \mathcal{N}$ represents an associated object instance $O_i$ and stores its persistent identifier, semantic label, centroid, axis-aligned bounding box (AABB), confidence score, and other geometric attributes. As new observations become available, node attributes are updated while preserving the persistent identifier.
Optionally, descriptive captions can be attached by selecting informative object views and applying a lightweight vision-language captioning model~\cite{gu2024conceptgraphs}. Since caption generation is decoupled from graph construction, it can be performed asynchronously without affecting the online mapping pipeline.

\subsubsection{Edges.}
Edges encode spatial or semantic relationships between graph nodes. We distinguish two complementary relation types: (i) online spatial relations inferred directly from object geometry, and (ii) optional open-set semantic relations inferred asynchronously using a \gls{vlm}. This separation enables efficient low-latency access to spatial relations for robot planning while allowing richer semantic knowledge to be incorporated without interrupting online mapping.

\paragraph{1) Online Spatial Relations:} 
Spatial relations are inferred online using simple geometric tests based on object centroids and AABBs. We consider relations such as \emph{near} (centroid distance below a threshold), \emph{on}/\emph{under} (sufficient horizontal overlap with limited vertical separation), \emph{inside} (bounding-box containment), and relative ordering relations 
derived from centroid offsets in the global coordinate frame. Each spatial relation is assigned a confidence score computed from normalized geometric measures. 

\paragraph{2) Open-set Semantic Relations:} 
To capture relations that cannot be reliably inferred from geometry alone, \coolname{} optionally enriches the scene graph using a \gls{vlm}. This process is independent of object extraction and graph construction, and is applied only to augment the graph with open-set semantic relations. 
Specifically, we render a set of keyframes following~\cite{ha2024rgbd} from the reconstructed scene with persistent instance identifiers overlaid near visible objects, and query the \gls{vlm} to predict directed relation triplets $(i, r, j)$. 
Predictions from individual keyframes are aggregated through multi-view voting. For each object pair, edge confidence is computed as the ratio of positive relation predictions to the number of keyframes in which both objects are co-visible. Relations with sufficient multi-view support are retained, reducing viewpoint-dependent ambiguity while improving the robustness of semantic relation prediction.

\section{Experiments}
\label{sec:exp}

We evaluate \coolname{} from three complementary perspectives corresponding to the proposed dual-representation architecture. First, we assess the reliability-aware semantic field through open-vocabulary 3D semantic segmentation (Sec.~\ref{subsec:exp:segmentation}). Next, we evaluate the persistent scene graph on 3D object grounding from relational natural language queries (Sec.~\ref{subsec:exp:grounding}). Finally, we demonstrate the practical utility of the complete system on a mobile robotic platform in real-world environments (Sec.~\ref{subsec:exp:robot}). Ablation studies further examine the contributions of individual components (Sec.~\ref{subsec:exp:ablation}).

\subsection{Implementation Details}
\label{subsec:exp:implementation}

We summarize the key implementation details of the proposed framework.
For 2D feature extraction, SAM~\cite{kirillov2023segment} is used on benchmark datasets, while MobileSAMv2~\cite{zhang2023mobilesamv2} is adopted for real-world robotic experiments to enable real-time operation.
The resulting segmentation masks are then encoded using the OpenCLIP ViT-B/16 model to obtain language-aligned semantic embeddings.
For query-conditioned object extraction, DBSCAN clustering is performed with radius $r=\beta \cdot \texttt{voxel\_size}$, where $\texttt{voxel\_size}=0.05\,\mathrm{m}$ and $\beta=1.5$. 
We set $k_{\min}=5$ and discard clusters containing fewer than $\texttt{min\_voxels}=10$ voxels. 
The semantic residual $r(v)$ is updated with decay factor $\alpha=0.9$.
Reliability-aware confidence estimation uses $\lambda=0.6$ and $\kappa=5.0$. 
Object association employs label-aware centroid matching with
a distance threshold of $\tau=0.35\,\mathrm{m}$. 
When open-set semantic relations are enabled, GPT-4o-mini predicts relation triplets, which are retained with at least three votes across up to 30 keyframes.

\subsection{3D Semantic Segmentation}
\label{subsec:exp:segmentation}

\subsubsection{Settings.}
\emph{(i) Task:}
We evaluate the quality of the proposed reliability-aware semantic field through open-vocabulary 3D semantic segmentation. Given a language query, object instances are extracted from the dense semantic field and compared against ground-truth semantic annotations. 
\emph{(ii) Baselines:} 
We compare our approach against recent \gls{gs}-based open-vocabulary methods, including LangSplat~\cite{qin2024langsplat}, OpenGaussian~\cite{wu2024opengaussian}, and InstanceGaussian~\cite{li2025instancegaussian}.
In addition, point cloud-based methods, including ConceptFusion~\cite{jatavallabhula2023conceptfusion} and OpenGS-Fusion~\cite{yang2025opengs}, are compared to assess performance under online mapping settings.
For all baselines, we use the official hyperparameters or tune them on validation scenes for fair comparison.
\emph{(iii) Metrics:} 
Experiments are conducted on ScanNet~\cite{dai2017scannet} and Replica~\cite{straub2019replica} with ground-truth 3D semantic annotations.
Performance is reported using \textit{mIoU} and \textit{mAcc}.
Following common practice, Replica's 88 semantic categories are remapped to the ScanNet20~\cite{dai2017scannet} and NYUv2-40~\cite{silberman2012indoor} label sets to ensure consistent evaluation.

\subsubsection{Results.}

\begin{table}[t]
\centering
\caption{
Quantitative comparison of 3D semantic segmentation on ScanNet and Replica datasets.
(Best and second-best results are highlighted in \textbf{bold} and \underline{underlined}.)}
\label{tab:segmentation}
\renewcommand{\arraystretch}{1.15}
\resizebox{\textwidth}{!}{%
\begin{tabular}{c l|ccccc|ccccc}
\hline
\multirow{2}{*}{Mode} & \multirow{2}{*}{Method} 
& \multicolumn{5}{c|}{ScanNet} 
& \multicolumn{5}{c}{Replica} \\
\cline{3-12}
& 
& \multicolumn{2}{c}{ScanNet20} 
& \multicolumn{2}{c}{NYUv2-40} 
& 
& \multicolumn{2}{c}{ScanNet20} 
& \multicolumn{2}{c}{NYUv2-40} 
& \\
& 
& mIoU$\uparrow$ & mAcc$\uparrow$ & mIoU$\uparrow$ & mAcc$\uparrow$ & FPS$\uparrow$
& mIoU$\uparrow$ & mAcc$\uparrow$ & mIoU$\uparrow$ & mAcc$\uparrow$ & FPS$\uparrow$\\
\hline
\multirow{3}{*}{Offline}
& LangSplat~\cite{qin2024langsplat}  & 3.78   & 9.11   & 3.46   & 8.43  & --  & 3.84 & 8.75    & 3.56    & 8.26        & -- \\
& OpenGaussian~\cite{wu2024opengaussian}  & 30.08 & 45.59 & 27.53 & 42.17 & -- & 25.63 & 36.24  & 23.45  & 31.59   & -- \\
& InstanceGaussian~\cite{li2025instancegaussian} & \textbf{40.32} & \textbf{55.78} & \textbf{36.88} & \textbf{52.91} & -- & 32.24 & \underline{45.99}  & 29.88  & 41.01   & -- \\

\hline
\multirow{3}{*}{Online}
& ConceptFusion~\cite{jatavallabhula2023conceptfusion}    & 12.80    & 22.62    & 11.72    & 20.93    & 0.52  & 13.05    & 29.89    & 11.51    & 27.66    & 0.49 \\
& OpenGS-Fusion~\cite{yang2025opengs}    & 27.09 & 39.62 & 24.69 & 35.95 & \textbf{1.22}  & \underline{34.84} & 45.15    & \underline{31.98}   & \underline{41.93}    &  \textbf{1.02}\\
& \textbf{\coolname{} (Ours)}    
& \underline{36.75} & \underline{53.82}  & \underline{33.52} & \underline{50.64} &  \underline{1.20}
& \textbf{35.37} & \textbf{47.67} & \textbf{32.87} & \textbf{42.85} & \textbf{1.02}  \\
\hline
\end{tabular}%
}
\end{table}

\begin{figure}[t]
	\centering
	\includegraphics[width=0.9\columnwidth]{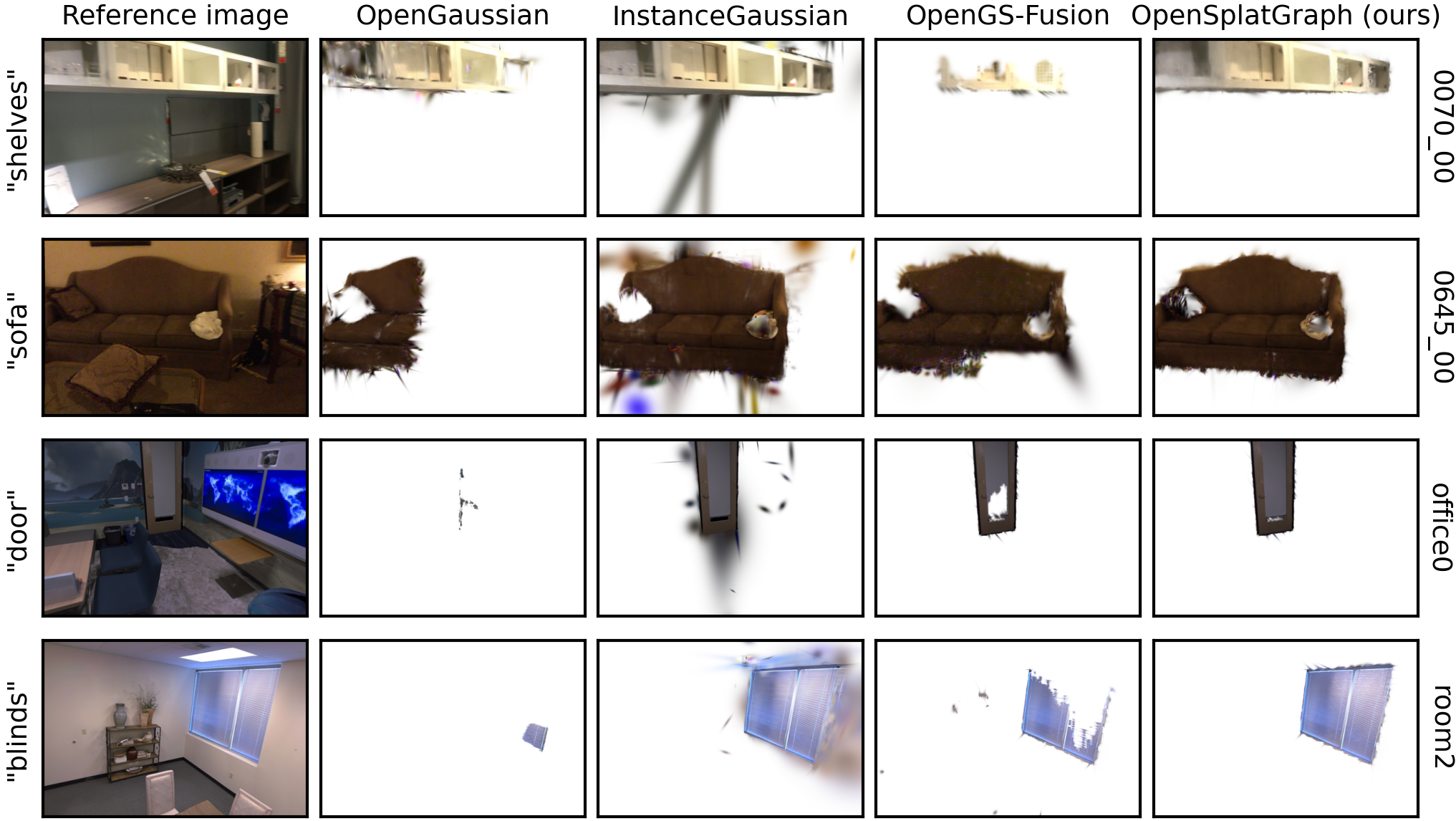}
	\centering
	\caption{Qualitative results of 3D semantic segmentation on ScanNet and Replica datasets.}
	\label{fig:segmentation}
\end{figure} 

Table~\ref{tab:segmentation} reports quantitative results for open-vocabulary 3D semantic segmentation on ScanNet and Replica. Among the \emph{offline} methods, InstanceGaussian achieves the best performance, benefiting from its bottom-up instance aggregation strategy~\cite{li2025instancegaussian}. Although \coolname{} is designed for online semantic mapping rather than offline segmentation, it achieves competitive performance with InstanceGaussian while consistently outperforming other Gaussian-based methods, including LangSplat and OpenGaussian. More importantly, under the \emph{online} setting, \coolname{} consistently surpasses ConceptFusion and OpenGS-Fusion on both mIoU and mAcc, demonstrating that the proposed reliability-aware semantic field produces more accurate and geometrically coherent object instances.

Figure~\ref{fig:segmentation} presents qualitative comparisons on both datasets. Compared with prior methods, \coolname{} produces more spatially coherent object instances with better alignment to object boundaries, particularly for thin structures and partially observed objects such as \emph{``shelves''}, \emph{``door''}, and \emph{``blinds''}. Modeling observation reliability effectively suppresses fragmented and noisy predictions. These qualitative observations are consistent with the quantitative results and demonstrate that the proposed reliability-aware semantic field provides a stronger foundation for persistent scene graph construction.

\subsection{3D Object Grounding}
\label{subsec:exp:grounding}

\subsubsection{Settings.} 
\emph{(i) Task:} We evaluate the persistent scene graph through 3D object grounding under complex natural language queries. Following the ConceptGraphs protocol~\cite{gu2024conceptgraphs},  we consider three types of queries: \emph{a) descriptive} (e.g., ``a pillow on top of the sofa''), \emph{b) affordance} (e.g., ``something to tell the time''), and \emph{c) negation} (e.g., ``something to sit on other than a chair''). 
\emph{(ii) Baselines:} We compare against ConceptGraphs~\cite{gu2024conceptgraphs}, the only existing open-vocabulary 3D scene graph method with a publicly available grounding benchmark and evaluation protocol on Replica~\cite{straub2019replica}. We adopt its query set and protocol for direct comparison, evaluating both CLIP retrieval based on query--embedding similarity and LLM retrieval based on structured node descriptions.
\emph{(iii) Metrics:} Performance is evaluated using the standard Recall@$k$ (R@$k$) metric~\cite{wald2020learning} following the ConceptGraphs protocol.

\subsubsection{Results.}

\begin{table*}[t]
\centering
\caption{Quantitative comparison of 3D object grounding on
Replica across different query types and retrieval strategies.
(ConceptGraphs results are reported from~\cite{gu2024conceptgraphs}.)
}
\label{tab:grounding}
\setlength{\tabcolsep}{4pt}
\renewcommand{\arraystretch}{1.15}
\resizebox{\textwidth}{!}{%
\begin{tabular}{l|ccc|ccc|ccc|ccc}
\hline
\multirow{3}{*}{Query Type}
& \multicolumn{6}{c|}{CLIP Retrieval}
& \multicolumn{6}{c}{LLM Retrieval} \\
\cline{2-13}
& \multicolumn{3}{c|}{ConceptGraphs}
& \multicolumn{3}{c|}{\coolname{}}
& \multicolumn{3}{c|}{ConceptGraphs}
& \multicolumn{3}{c}{\coolname{}} \\
\cline{2-13}
& R@1 & R@2 & R@3
& R@1 & R@2 & R@3
& R@1 & R@2 & R@3
& R@1 & R@2 & R@3 \\
\hline
Descriptive
& \textbf{0.59} & 0.82 & 0.86
& 0.45 & 0.82 & \textbf{0.92}
& \textbf{0.61} & \textbf{0.64} & 0.64
& 0.50 & 0.59 & \textbf{0.69} \\
Affordance
& 0.43 & 0.57 & 0.63
& \textbf{0.54} & \textbf{0.61} & \textbf{0.79}
& 0.57 & 0.63 & 0.66
& \textbf{0.61} & \textbf{0.71} & \textbf{0.75} \\
Negation
& 0.26 & 0.60 & 0.71
& \textbf{0.51} & \textbf{0.64} & \textbf{0.82}
& 0.80 & 0.89 & \textbf{0.97}
& 0.80 & 0.89 & 0.94 \\
\hline
\end{tabular}%
}
\end{table*}

\begin{figure}[t]
	\centering
	\includegraphics[width=0.8\columnwidth]{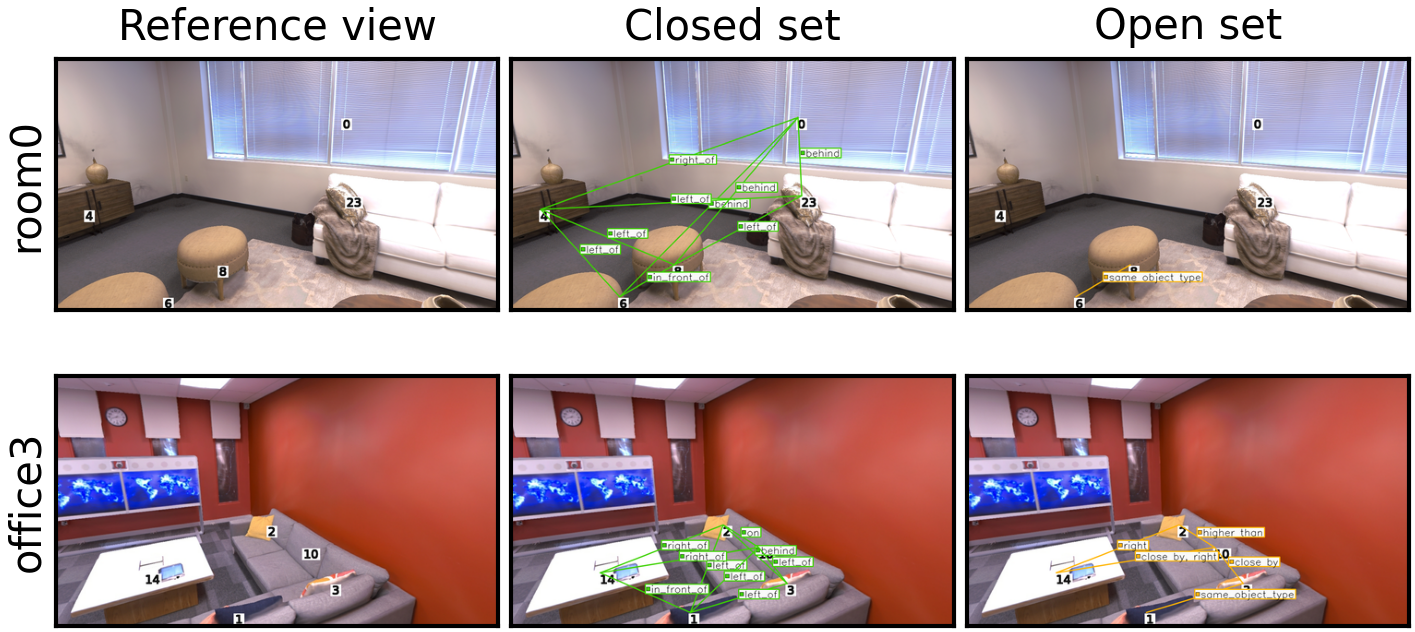}
	\centering
	\caption{
    Qualitative visualization of open-vocabulary 3D scene graph construction on the Replica dataset. For visualization, asymmetric relations are displayed from nodes with lower instance IDs to those with higher IDs.
    }
	\label{fig:graph}
\end{figure} 

Table~\ref{tab:grounding} reports quantitative results for 3D object grounding on the Replica dataset across three query categories and two retrieval strategies. Overall, \coolname{} achieves competitive grounding performance and consistently improves Recall@$k$ over ConceptGraphs in several retrieval settings. The largest improvements are observed for the \emph{affordance} and \emph{negation} queries, which require reasoning beyond direct semantic similarity. These gains demonstrate the benefit of maintaining a persistent object-centric scene graph, where explicit object identities and inter-object relations provide complementary information to the underlying dense semantic field during language grounding. 

Figure~\ref{fig:graph} visualizes representative scene graphs predicted by \coolname{} on two Replica scenes. For clarity, the 3D scene graphs are projected into 2D graph visualizations from a reference viewpoint. The predicted object relationships are spatially and semantically consistent with the underlying scene geometry. In addition, open-set relation prediction enables diverse semantic predicates, such as ``same object type'', that cannot be predefined by fixed relation vocabularies, further enriching the scene graph for downstream reasoning. 
Additional direct scene-graph evaluation is provided in the supplementary material.

\subsection{Real-world Robotic Experiments}
\label{subsec:exp:robot}
We evaluate the practical utility of \coolname{} in a real-world indoor environment using a Stretch~3 (Hello Robot) mobile manipulator equipped with an Intel RealSense D435i RGB-D camera and an arm with a gripper. 
During exploration, the robot incrementally scans the environment to acquire RGB-D observations while constructing an online Gaussian semantic map. Following OpenGS-Fusion~\cite{yang2025opengs}, camera poses are estimated using ORB-SLAM3~\cite{campos2021orb}. The proposed reliability-aware semantic field and persistent scene graph are then built online from the fused semantic map to support language-guided robotic perception. 

\begin{figure}[t]
	\centering
	\includegraphics[width=1.0\columnwidth]{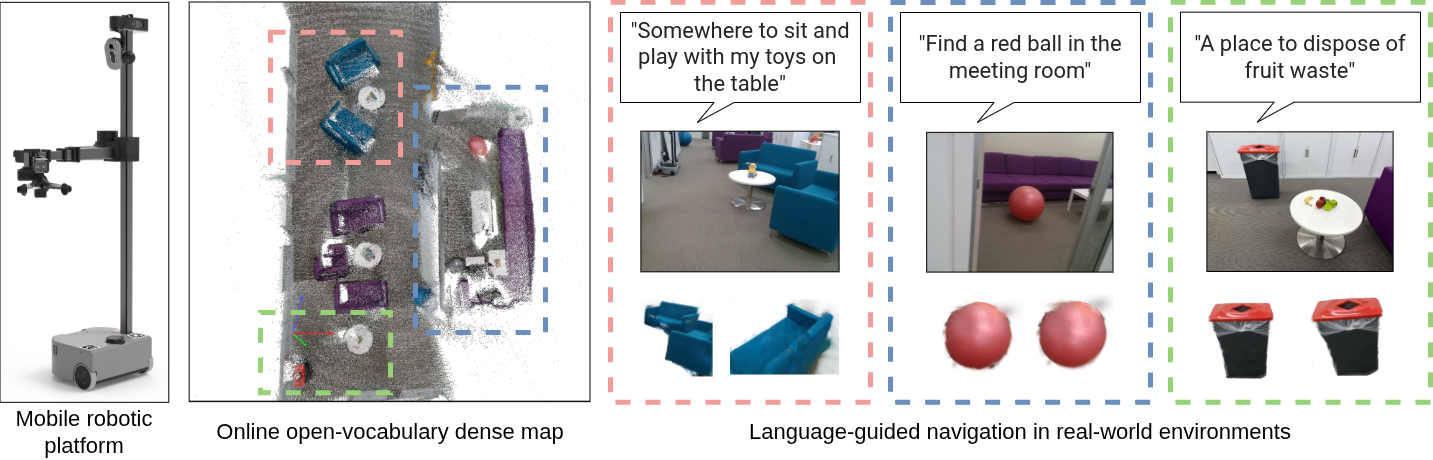}
	\centering
	\caption{Real-world robotic navigation experiments with \coolname{}.
The framework constructs an open-vocabulary semantic map online and supports language-guided navigation in indoor environments. For each query, we visualize the rendered target object from the reference view and from an additional arbitrary viewpoint.}
	\label{fig:navigation}
\end{figure} 

To demonstrate the effectiveness of the proposed perception framework for downstream robotics, we evaluate two representative tasks, as illustrated in Fig.~\ref{fig:overview} and Fig.~\ref{fig:navigation}. 
In a \emph{navigation} task, the robot receives abstract natural language queries that may include semantic, relational, or functional descriptions. Using the persistent scene graph, the queried object is grounded in the reconstructed environment, after which the robot plans and executes navigation toward the target location. 
In a \emph{manipulation} task, the robot performs language-guided pick-and-place by grounding an object query (e.g., ``teddy bear'') within the persistent scene graph, localizing the corresponding object in the dense semantic map, and executing grasping and placement into a designated container.
These experiments illustrate that \coolname{} provides a unified structured and open-vocabulary perception backbone that supports closed-loop robotic navigation and manipulation in real-world environments.
A demonstration video of the complete real-world experiments is included in the supplementary material.

\subsection{Ablation Study}
\label{subsec:exp:ablation}

\begin{table}[t]
\centering
\caption{Progressive ablation of OpenSplatGraph components.}
\label{tab:ablation}
\begin{tabular}{lccc}
\toprule
Configuration & mIoU$\uparrow$ & mAcc$\uparrow$ & R@1$\uparrow$ \\
\midrule
Baseline & 30.97 & 42.39 & -- \\
+ Reliability-aware semantic modeling & 36.06 & 50.74 & -- \\
+ Persistent object association & 36.06 & 50.74 & 0.36 \\
+ Scene graph reasoning (Full) & \textbf{36.06} & \textbf{50.74} & \textbf{0.50} \\
\bottomrule
\end{tabular}
\end{table}

Table~\ref{tab:ablation} presents a progressive ablation of the proposed framework. 
Starting from a baseline model, introducing reliability-aware semantic modeling significantly improves segmentation performance, demonstrating that modeling observation quality leads to more accurate and coherent object instances.
Adding persistent object identities does not affect segmentation metrics but improves object grounding by enabling consistent object association across multiple queries.
In this setting, object grounding relies solely on semantic similarity without exploiting relational information.
Finally, incorporating scene graph reasoning yields the largest gain in relational grounding, highlighting the benefit of explicit inter-object modeling.
Overall, the full framework achieves the best performance across all metrics, validating the effectiveness of the proposed design.

\subsection{Limitations}
\label{subsec:limitations}

While \coolname{} enables persistent scene graph construction from an online open-vocabulary semantic map, several limitations remain. First, object extraction is query-conditioned, and the framework does not maintain a complete query-independent object inventory. Instead, object instances are accumulated incrementally as new queries are issued. Second, object association relies on lightweight semantic and geometric cues, which may become less reliable in highly cluttered scenes or under severe viewpoint changes. Finally, open-set semantic relations are inferred using an external \gls{vlm}, introducing additional computational overhead and dependence on the model's reasoning capability. Addressing these limitations through more robust object association and query-independent proposal mechanisms is a promising direction for future work.

\section{Conclusion}
\label{sec:conclusion}

We presented \coolname{}, a unified framework that jointly maintains a reliability-aware open-vocabulary semantic field and a persistent 3D scene graph for online robotic perception. The proposed semantic field enables robust query-conditioned object extraction from dense Gaussian-based maps, while the persistent scene graph incrementally accumulates object-centric knowledge to support structured relational reasoning. Extensive experiments on standard benchmarks demonstrate competitive performance in open-vocabulary semantic segmentation and 3D object grounding, and real-world robotic experiments validate the effectiveness of the proposed framework for language-guided navigation and manipulation. Future work will address the discussed limitations and explore integration with multi-step task planning in more complex environments.

\subsubsection*{Acknowledgements} 
This work was supported in part by the Australian Research Council Discovery Project under Grant DP250103634, and in part by the Commonwealth Scientific and Industrial Research Organisation (CSIRO). The authors acknowledge continued support from the CSIRO’s Embodied AI Cluster.

\bibliographystyle{splncs04}
\bibliography{main}

\clearpage
\pagestyle{empty}
\appendix

\begin{center}
{\Large\bfseries
OpenSplatGraph: \\
From Dense Semantic Maps to Structured Scene Graphs for Open-Vocabulary Robot Perception\par}
\vspace{1em}
{\large\bfseries Supplementary Material\par}
\end{center}
\vspace{1em}

\section{Additional 3D Semantic Segmentation Results}
\label{supp:segmentation}

We provide additional comparisons with GaussianGraph~\cite{wang2025gaussiangraph} and DynamicGSG~\cite{ge2025dynamicgsg} for open-vocabulary 3D semantic segmentation on ScanNet and Replica. Table~\ref{tab:supp_segmentation} reports the results alongside \coolname{}.
Compared with GaussianGraph, \coolname{} achieves higher mIoU on both datasets. DynamicGSG uses a different class and evaluation mapping; therefore, its Replica results are included for reference rather than direct comparison.

\begin{table}[ht]
\centering
\caption{Additional open-vocabulary 3D semantic segmentation results on ScanNet and Replica. 
$^\dagger$DynamicGSG uses a different evaluation mapping.
}
\label{tab:supp_segmentation}
\renewcommand{\arraystretch}{1.15}
\setlength{\tabcolsep}{5pt}
\begin{tabular}{l|cc|cc}
\hline
\multirow{2}{*}{Method}
& \multicolumn{2}{c|}{ScanNet}
& \multicolumn{2}{c}{Replica} \\
& mIoU$\uparrow$ & mAcc$\uparrow$
& mIoU$\uparrow$ & mAcc$\uparrow$ \\
\hline
GaussianGraph~\cite{wang2025gaussiangraph}
& 31.09 & 48.91 & 31.18 & 49.14 \\
DynamicGSG~\cite{ge2025dynamicgsg}
& -- & -- & 31.06$^\dagger$ & 54.04$^\dagger$ \\
\coolname{} (Ours)
& \textbf{36.75} & \textbf{53.82} & \textbf{35.37} & 47.67 \\
\hline
\end{tabular}
\end{table}

\section{Direct Scene-Graph Evaluation}
\label{supp:graph_eval}

To directly evaluate graph correctness and identity persistence, we compare \coolname{} with ConceptGraphs~\cite{gu2024conceptgraphs}.
Following ConceptGraphs, we report node precision for object instances, edge precision for predicted inter-object relations, and duplicate-node rate to assess identity consistency across sequential observations.

\begin{table}[ht]
\centering
\caption{Direct scene-graph evaluation against ConceptGraphs.
}
\label{tab:supp_graph_eval}
\renewcommand{\arraystretch}{1.15}
\setlength{\tabcolsep}{5pt}
\begin{tabular}{l|ccc}
\hline
Method
& Node Prec.$\uparrow$
& Edge Prec.$\uparrow$
& Dup. Rate (\%)$\downarrow$ \\
\hline
ConceptGraphs~\cite{gu2024conceptgraphs}
& 0.71 & 0.88 & 6.77 \\
\coolname{} (Ours)
& \textbf{0.73} & \textbf{0.91} & \textbf{5.94} \\
\hline
\end{tabular}%
\end{table}

As shown in Table~\ref{tab:supp_graph_eval}, \coolname{} achieves higher node and edge precision while reducing the duplicate-node rate compared to ConceptGraphs.
These results complement the object grounding evaluation by directly assessing graph correctness and identity consistency.

\end{document}